\documentclass[letterpaper]{article}
\usepackage[preprint]{reco}
\usepackage[hyphens]{url}
\usepackage{graphicx}
\usepackage{natbib}
\usepackage{caption}
\usepackage{algorithm}
\usepackage{algorithmic}

\usepackage{amsmath}

\usepackage[table]{xcolor}
\usepackage{graphicx}
\usepackage{subcaption}
\usepackage{multirow}

\usepackage{newfloat}
\usepackage{listings}
\DeclareCaptionStyle{ruled}{labelfont=normalfont,labelsep=colon,strut=off}
\floatstyle{ruled}  
\newfloat{listing}{tb}{lst}{}
\floatname{listing}{Listing}
\usepackage[table]{xcolor}
\definecolor{recoblue}{RGB}{233,238,250}
\definecolor{modelyellow}{RGB}{255,245,200}
\usepackage{pifont}
\usepackage{booktabs}
\usepackage{tabularx}

\title{Fewer Tokens, Smaller Cache: Reward-Coordinated Efficient Reasoning}
\author{
    Qiyuan Zhu\textsuperscript{\rm 1}\equalcontrib,
    Dezhi Li\textsuperscript{\rm 1}\equalcontrib,
    Pengyu Cheng\textsuperscript{\rm 2}\equalcontrib,
    Tianle Chen\textsuperscript{\rm 2},
    Jiacheng Wang\textsuperscript{\rm 2},
    Ruijie Shen\textsuperscript{\rm 3},\\
    Hao Gu\textsuperscript{\rm 1},
    Sida Lin\textsuperscript{\rm 1},
    Zirui Liu\textsuperscript{\rm 4},
    Jiacheng Liu\textsuperscript{\rm 1}\corresponding,
    Sirui Han\textsuperscript{\rm 1}\corresponding
}
\affiliations{
    \textsuperscript{\rm 1}HKUST,
    \textsuperscript{\rm 2}XJTU,
    \textsuperscript{\rm 3}TSE,
    \textsuperscript{\rm 4}PKU
}

\begin{document}

\maketitle
\begin{abstract}
Large Reasoning Models (LRMs) excel on complex tasks through long chain-of-thought (CoT)
reasoning, but their lengthy intermediate steps cause severe overthinking that inflates
inference cost. KV-cache compression is a common solution, yet existing reasoning-oriented
methods apply a uniform policy across the trajectory and judge compression only by what it
removes from the cache. Two observations point the other way. First, a reasoning state's
tolerance to context loss varies along the trajectory, and process reward tracks it: deleting tokens at high-reward steps
preserves accuracy far better than deleting the same budget at random. Second, compression is not free on the
generation side, since a smaller cache leads the model to generate more tokens, partly canceling
the saving. Together these motivate coordinating both sides under a single process reward. We
propose ReCo (Reward-Coordinated Compression), a step-wise framework in which a lightweight
process-reward estimator scores each completed step and drives three components: (1)
reward-adaptive KV-cache compression that shrinks the retained cache harder at high-reward
steps and less at low-reward ones, (2) a reward-banded penalty on reflection tokens that curbs redundant
generation, and (3) confidence-based early stopping that triggers when the reasoning is reliable.
Across three reasoning models and six benchmarks, ReCo reduces generated tokens by
$37\%$--$65\%$ and end-to-end latency by $2.08\times$--$2.35\times$ over Full CoT,
all while largely preserving accuracy.
\end{abstract}

\section{Introduction}
\begin{figure}[t!]
    \centering
    \includegraphics[width=\linewidth]{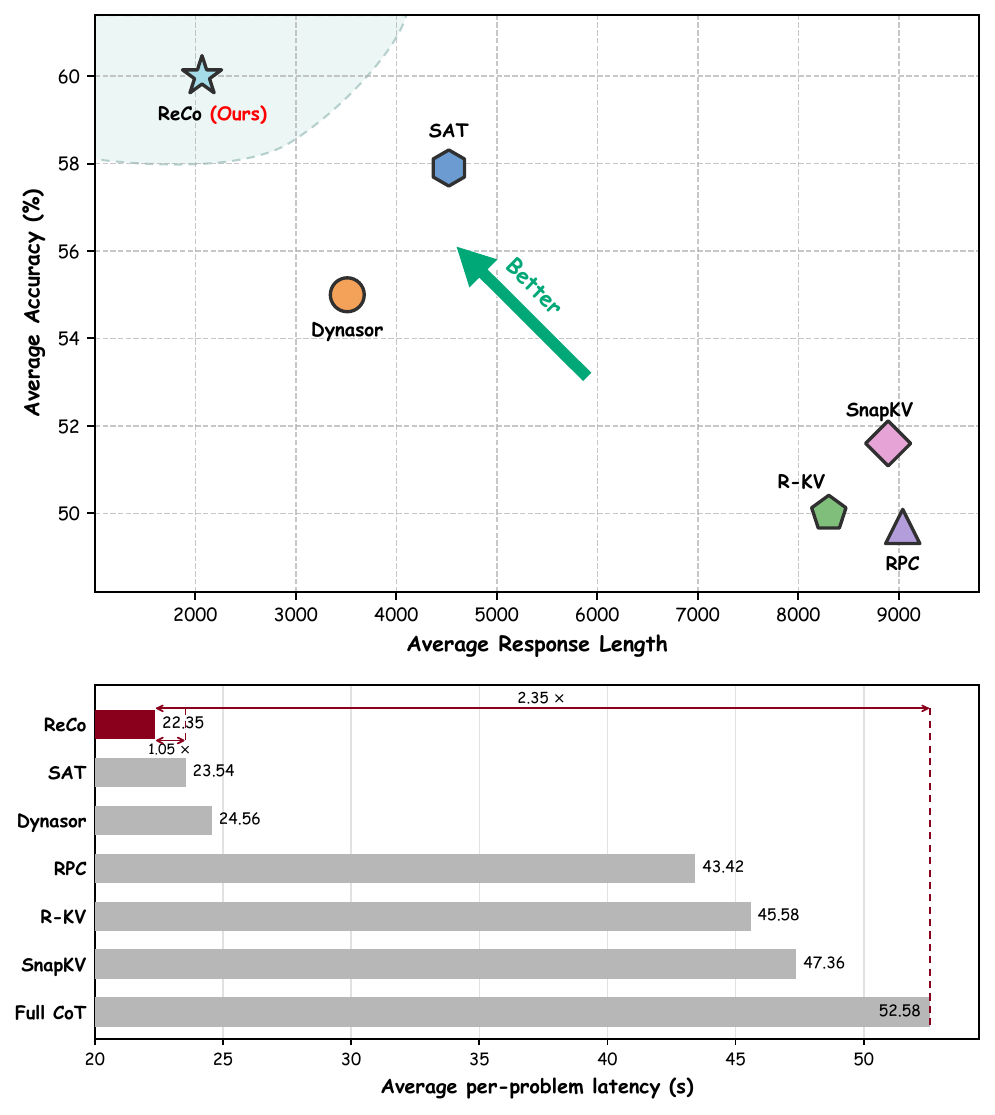}
    \caption{
    Introduction overview on DeepSeek-R1-Distill-Qwen-7B.
    Top: accuracy vs.\ average response length; ReCo is the most accurate compressed method while generating far fewer tokens.
    Bottom: per-problem latency; ReCo is the fastest overall, cutting latency by $2.35\times$ over Full CoT.
    }
    \label{fig:intro_teaser}
\end{figure}
Large Reasoning Models (LRMs), such as OpenAI-o1~\cite{jaech2024openai}, DeepSeek-R1~\cite{guo2025deepseek}, and Gemini~\cite{comanici2025gemini}, have recently achieved remarkable progress across a wide range of domains, demonstrating strong capabilities on complex tasks such as mathematical reasoning, code generation, and scientific problem solving~\cite{zhu2026outlier,xu2026nanoresearch, cao2025towards, cao2026pushing, gu2026qarl}. However, such capabilities are largely driven by long chain-of-thought (CoT) reasoning, where the model produces lengthy intermediate steps before arriving at a final answer. While effective on hard problems, this paradigm frequently causes models to \emph{overthink}~\cite{chen2024not}: they generate excessively long reasoning chains even for simple inputs, incurring token overhead, increased latency, and high computational cost.

\begin{figure*}[t]
  \centering
  \includegraphics[width=\textwidth]{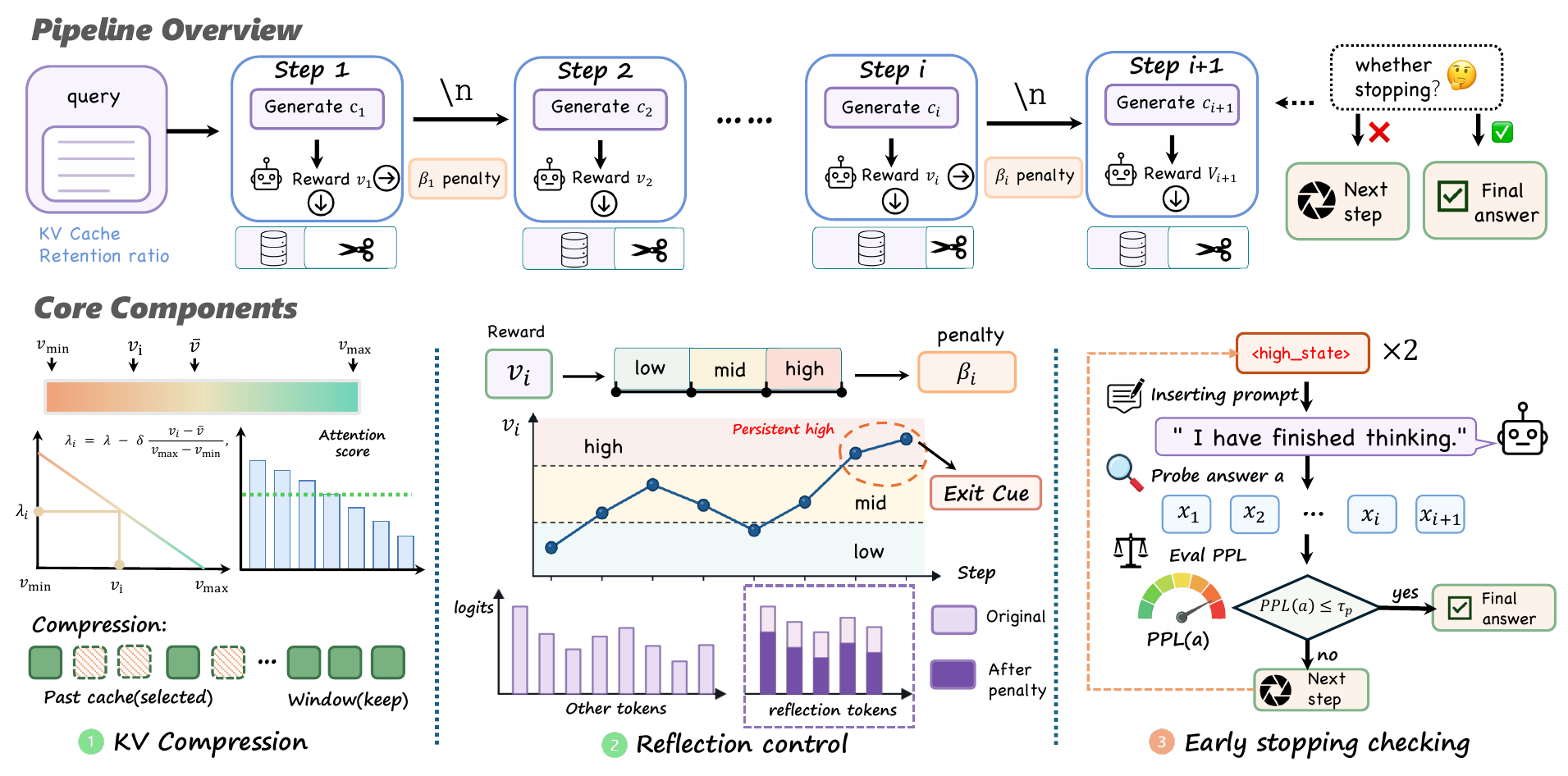}
\caption{Overview of \textbf{ReCo}. After each step $c_i$, a process reward $v_i$ drives three
  components: \textbf{\ding{182}~KV Compression} sets a reward-modulated retention ratio,
  shrinking the cache harder at high-reward steps; \textbf{\ding{183}~Reflection Control} maps $v_i$ to a
  reward-banded logit penalty $\beta_i$ on reflection tokens; and \textbf{\ding{184}~Early Stopping} halts
  once answer perplexity satisfies $\mathrm{PPL}(a)\le\tau_p$.}
  \label{fig:overview}
\end{figure*}

A common strategy for mitigating these inference costs is KV-cache compression, since the key-value (KV) cache accumulated during autoregressive decoding dominates both memory footprint and per-token attention cost~\cite{li2024survey, shi2024keep, hu2026hierarchical}. Classical methods evict cache entries deemed unimportant by attention scores (e.g., SnapKV~\cite{li2024snapkv} and PyramidKV~\cite{cai2024pyramidkv}), merge redundant entries into representative ones~\cite{zhang2024cam}, or quantize them into low-bit formats~\cite{hooper2024kvquant}. These methods, however, were designed primarily for long-context settings, where the cache is dominated by a long \emph{static} prompt that can be pruned once after prefilling; in reasoning, by contrast, the cache is dominated by the model's own \emph{growing} chain of thought, whose content must keep supporting the very generation that produces it. Recent work has begun to adapt KV compression to this reasoning setting~\cite{cai2026r, song2026reasoning}, yet it still applies a \emph{uniform} compression policy across the entire trajectory, ignoring that reasoning steps differ in how much they can afford to be compressed. More fundamentally, they weigh compression only by what it removes from the cache, and leave unexamined whether discarding context affects the reasoning the model has yet to generate.

A closer look at these gaps reveals two observations. \emph{First, compression tolerance varies
along the reasoning trajectory.} A process reward, which scores how promising and
on-track the current path is, reliably tracks it: at high-scoring steps the reasoning
state is on-track and withstands aggressive pruning of its context, whereas
a low-scoring state is still exploring and fragile, so deleting tokens from high-reward
steps consistently outperforms deleting the same budget at random
(Sec.~\ref{subsec:motivation_reward}). \emph{Second, the saving from KV compression is
partly offset by longer generation}: at a fixed compression rate the average output length
exceeds the full-cache baseline, and on MATH-500 up to $79.8\%$ of problems generate more
tokens once the cache is compressed (Sec.~\ref{subsec:motivation_length}). Together these show that, on
reasoning models, \textbf{compression must respect how the reasoning trajectory varies from
step to step, and it cannot be decoupled from generation length: shrinking the cache while
the output grows can erode the savings.}

Concretely, we propose \textbf{ReCo}
(\textbf{Re}ward-\textbf{Co}ordinated Compression), a framework
that improves KV-cache compression and resolves its length inflation on reasoning models
by coordinating it with generation control (Figure~\ref{fig:overview}). As
each step completes, a lightweight $30$M process-reward
estimator assigns it a scalar reward reflecting how promising and
on-track the current reasoning path is, and this single reward drives three
coordinated components.
\emph{(1) KV-cache compression} maps the reward to a
step-dependent retention ratio for the accumulated cache, shrinking
the whole cache harder when the latest step scores high and less when it scores low,
replacing the uniform policy of prior reasoning-oriented KV methods.
\emph{(2) Reflection control} maps the same reward to a
reward-banded logit penalty on self-reflection tokens that steers the model toward more concise reasoning
when the step scores highly, curbing the length inflation that would otherwise erode the
compression saving. \emph{(3) Early stopping}, triggered by persistently high reward, terminates the
chain of thought once an answer-confidence probe judges the model ready to commit.
Because all three share the same per-step reward, they stay
consistent, spending compute where reasoning is still productive and
withholding it where the model has already settled.

Across three reasoning models and six benchmarks, ReCo cuts the average generated tokens
by $37\%$--$65\%$ and lowers end-to-end latency by $2.08\times$--$2.35\times$ over full-cache
CoT, with minimal accuracy loss. Under matched settings, it
attains the smallest accuracy drop among all compressed methods: cache-only baselines lose far
more accuracy and, consistent with the length-inflation effect above, often generate
\emph{more} tokens than the full-cache model, whereas length-only baselines reach a comparable
speedup only by sacrificing accuracy.

\begin{figure*}[t!]
\centering

\includegraphics[width=\textwidth]{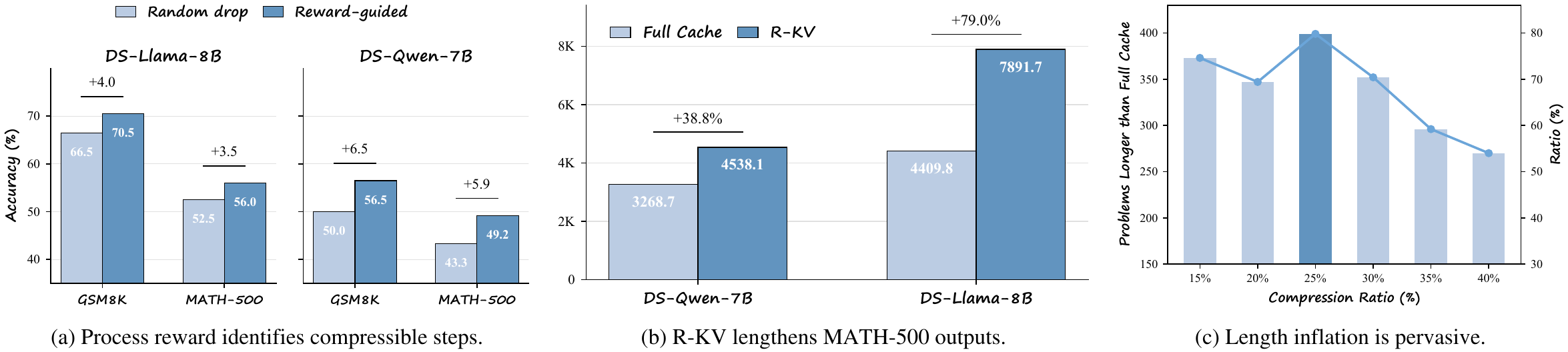}

\caption{
Motivating observations for reward-guided KV-cache compression and generation control.
(a) Deleting tokens from high-reward steps preserves accuracy better than
deleting the same budget from randomly chosen steps, indicating that process reward
tracks compression tolerance.
(b) Cache compression can backfire and lengthen generation on MATH-500.
(c) Over half of MATH-500 problems generate longer outputs once compressed,
and this holds across compression rates.
}
\label{fig:motivation_overview}
\end{figure*}

Our contributions are summarized as follows:
\begin{itemize}
    \item We argue that, on reasoning models, KV-cache compression is insufficient on its own and should be coupled with generation control: compressing the cache alone can backfire by lengthening the very reasoning it is meant to make cheaper, so the two are better governed jointly.
    \item We show that \emph{process reward} provides a reliable step-wise signal of compression tolerance, telling when the state can run on a smaller cache and when it needs more context, in contrast to the uniform policies of existing KV compression methods.
    \item We design a reward-coordinated, step-wise framework that unifies KV-cache compression, reflection control, and early stopping under one signal, reducing both token count and latency while preserving accuracy across three models and six benchmarks.
\end{itemize}

\section{Motivation and Observation}
\label{sec:motivation}

\subsection{Higher Process Reward, Higher Compression Tolerance}
\label{subsec:motivation_reward}

Reasoning steps leave the model in distinct states~\cite{chen2026adaptive}: at some the path is
already settled, at others it is still being worked out, so compression should adapt to the
state the trajectory is in rather than treat every step alike (steps split at newline
tokens, see Sec.~\ref{sec:prelim}). To test whether process reward captures this,
we score trajectories from DeepSeek-R1-Distill-Qwen-7B and DeepSeek-R1-Distill-Llama-8B on
GSM8K and MATH-500 with Skywork-o1-Open-PRM-7B~\cite{he_2024_16998085}, and compare two deletions that remove the \emph{same}
number of tokens: reward-guided, dropping $70\%$ of tokens within the top-$50\%$ high-reward
steps, versus random, removing the same amount from $50\%$ of randomly chosen steps.

As Figure~\ref{fig:motivation_overview}(a) shows, concentrating deletion on high-reward
steps consistently outperforms random step selection across both models and datasets, by
$4.0$/$3.5$ points on GSM8K/MATH-500 for Llama-8B and a larger $6.5$/$5.9$ for Qwen-7B. Since
both settings remove the same number of tokens, the only difference is the state of the steps
that bear the loss.

We attribute this to what the reward measures: how likely the current reasoning state lies on a
reliable path, not how indispensable its tokens are. High-reward steps are on-track and
consistent with their context, so removing tokens loses less information, whereas
low-reward steps are where the model is still exploring, so each token carries more information
and removing it more easily breaks the chain.
\textbf{Higher process reward thus marks a reasoning state that tolerates more disruption to its
context: the same removal loses less information and hence costs less accuracy.}
We call this its \emph{compression tolerance}: the reward decides \emph{how much} context the
current state can afford to lose. Here that loss is deleting tokens within steps; in ReCo it is a
smaller retained cache (Sec.~\ref{sec:kv}).

\subsection{KV Cache Compression Can Induce Longer Reasoning}
\label{subsec:motivation_length}

KV cache compression reduces inference cost by shrinking the retained context. For reasoning
models, however, this saving is not free: compressing the cache can lengthen the subsequent
reasoning, so the model may spend more tokens generating than it saves on the cache.

Figure~\ref{fig:motivation_overview}(b) illustrates this effect on MATH-500 under
R-KV~\cite{cai2026r} at a $25\%$ compression rate. Relative to the full cache, R-KV raises
the average generation length from $3268.7$ to $4538.1$ tokens ($+38.8\%$) on
DeepSeek-R1-Distill-Qwen-7B, and from $4409.8$ to $7891.7$ tokens ($+79.0\%$) on
DeepSeek-R1-Distill-Llama-8B, matching the R-KV rows of Table~\ref{tab:main}. Thus, while KV
compression lowers the per-token attention cost, the extra tokens it induces partly cancel
this saving.

\emph{Moreover, this is a pervasive per-problem phenomenon, not a few outliers.}
Figure~\ref{fig:motivation_overview}(c) reports the fraction of MATH-500 problems whose
compressed output exceeds the full-cache output across compression rates. The ratio peaks at
$79.8\%$ ($399/500$ problems) at a $25\%$ compression rate and stays above $50\%$ over the
whole $15$--$40\%$ range, confirming that length inflation is systematic. \textbf{On reasoning models, therefore, KV
compression should not be treated as a pure cache problem, but coupled with generation
control to account for its effect on output length, exactly what our method does.}

\section{Methodology}
\label{sec:methodology}

We propose \textbf{ReCo} (Figure~\ref{fig:overview}), which operates at the granularity of
reasoning steps: after each step, it scores the current reasoning state with a single lightweight reward and
uses that score to drive KV-cache compression (Sec.~\ref{sec:kv}), reflection control, and early
stopping (Sec.~\ref{sec:forward}).

\subsection{Preliminary}
\label{sec:prelim}

We begin by describing how the reasoning trajectory is partitioned into steps and how
the per-step reward signal is obtained, which together form the basis for all three
components of ReCo.

\paragraph{Reasoning steps as the unit of control.}
For a query $q$, the LRM $\mathcal{M}$ produces a reasoning trajectory
$\mathcal{C}$ before emitting the final answer. Following~\cite{huang2026sat},
we define reasoning steps by splitting $\mathcal{C}$ at newline tokens
(\texttt{\textbackslash n}), and write
$\mathcal{C}=\{c_1,\dots,c_T\}$, where
$c_i\sim\mathcal{M}(\cdot\mid q,c_{<i})$ denotes the autoregressive generation of the
$i$-th step. Since a completed step typically corresponds to a semantically coherent
reasoning unit, it provides a natural online decision point: given the partial
trajectory $c_{\le i}$, our framework determines how to compress the accumulated
KV cache and how to control the subsequent generation.

\paragraph{Reward signal.}
After each completed step $c_i$, we assign a scalar reward $v_i\in[0,1]$ to the current
reasoning trajectory $c_{\le i}$, estimating the likelihood of eventually reaching a
correct answer. For lightweight online, per-step use, we adopt
\textbf{Pilot}~\cite{huang2026sat}, a compact $30$M-parameter estimator distilled from
Skywork-o1-Open-PRM-7B, the same PRM behind our observation in
Sec.~\ref{subsec:motivation_reward}, giving $v_i = \mathrm{Pilot}(c_{\le i})$. Trained to
match this teacher, Pilot retains its step-level discriminability while staying light enough
to score every step online, so motivation and method share one reward family rather than two
unrelated scorers. This score is the \emph{unified signal} in ReCo that guides how the next step
$c_{i+1}$ is produced, coordinating both KV-cache compression and generation-side
control.

\subsection{Reward-Adaptive KV Compression}
\label{sec:kv}

This component compresses the accumulated KV cache periodically as reasoning proceeds,
under one guiding principle: \emph{the step reward decides how much context to keep}. As
established in Sec.~\ref{subsec:motivation_reward}, a high-reward state tolerates more
disruption to its context. Inspired by this, ReCo compresses while decoding, pruning more
aggressively when the reward is high and retaining more when it is low. We therefore let the
reward $v_i$ of the just-completed step set
the retention ratio $\lambda_i$ for the whole accumulated cache, refreshing this ratio at every
step rather than following a fixed, uniform schedule.

\paragraph{Reward-driven retention.}
Whenever the cache has grown by $S$ tokens since the last compression, i.e.,
$L - L_{i-1} \ge S$ with $L$ the current cache length and $L_{i-1}$ that after the
previous compression ($L_0$ being the prompt length), we compress the $L$ cached tokens
down to $m_i=\lambda_i L$. The ratio $\lambda_i$ is where the reward takes control: a
high reward marks a reliable reasoning state, so the accumulated context can tolerate more
aggressive pruning, while a low reward marks a step that is still exploring, where the
reasoning is fragile and more of the cache should be retained. We thus compare the latest
reward against the trajectory's own reward statistics, compressing harder when it is above
average and retaining more when below:
\begin{equation}
\lambda_i \;=\;
\underbrace{\lambda}_{\text{base ratio}}
\;-\;
\underbrace{\delta\,\frac{v_i-\bar v}{\,v_{\max}-v_{\min}\,}}_{\text{reward adjustment}},
\label{eq:budget}
\end{equation}
where $\bar v$, $v_{\min}$, $v_{\max}$ are the mean, minimum, and maximum step rewards in
the current trajectory, $\lambda$ is the base ratio, and $\delta$ bounds the adjustment
so that $\lambda_i\in[\lambda-\delta,\lambda+\delta]$. This normalization turns the raw
reward into a within-trajectory relative position, so $\lambda_i$ depends on whether a
step is high or low \emph{for that trajectory} rather than on its absolute reward value.
In effect, the latest step's reward sets a single retention level for the whole cache: when
the current reasoning state scores high, we shrink the accumulated cache more aggressively;
when it scores low, we retain more of it. This level is refreshed with the latest reward every
$S$ generated tokens at negligible overhead.

\paragraph{Attention-guided selection.}
Within the reward-allocated budget $m_i$, we retain the tokens most relevant to the
ongoing reasoning, scored by the attention they attract from the $w$ most recent tokens:
\begin{equation}
s_j \;=\; \sum_{t=L-w+1}^{L} \mathrm{softmax}_j\!\big(q_t^\top k_j\big),
\label{eq:imp}
\end{equation}
where $q_t$ and $k_j$ are the query and key of tokens $t$ and $j$. The $m_i$
highest-scoring tokens are kept, with the latest $w$ tokens always retained as the query
window; since $s_j$ is recomputed at each compression, tokens no longer attended to are
evicted automatically. This yields a clean division of labor: the reward decides
\emph{how much} survives each compression, and attention only decides \emph{what}
survives within that allowance.

\subsection{Reward-Adaptive Generation Control}
\label{sec:forward}

Compressing the cache alone is insufficient on its own: as shown in
Sec.~\ref{subsec:motivation_length}, compressing the cache tends to lengthen the subsequent
reasoning, and the added generation partly cancels the savings. ReCo therefore also governs the
tokens still to be generated, reusing the same per-step reward $v_i$ that drives compression.

\paragraph{Reward-banded reflection penalty.}
Long reasoning traces spend many tokens on explicit self-reflection, opening new branches
with reflection tokens such as ``Wait'', ``Hmm'', and ``Alternatively'', which drives much of the
overthinking on easy or already-settled
steps~\cite{wang2025wait, lotfi2026quantized}.
Rather than injecting instructions or fixing a global budget, we suppress this
\emph{directly at the decoding level}. We curate a set $\mathcal{R}$ of such reflection tokens (with their
tokenizer variants)
and penalize these reflection tokens during decoding, with the penalty strength set
by the just-completed step's reward, using two thresholds $\tau_{\ell}<\tau_{h}$ that split the reward into three
bands: no penalty in the lowest, a half penalty in the middle, and the full penalty
$\beta$ in the highest:
\begin{equation}
\beta_i \;=\;
\begin{cases}
0,          & v_i \le \tau_{\ell},\\[2pt]
\beta/2,    & \tau_{\ell} < v_i < \tau_{h},\\[2pt]
\beta,      & v_i \ge \tau_{h},
\end{cases}
\qquad \beta>0 .
\label{eq:beta_bands}
\end{equation}
While generating the $(i{+}1)$-th step, we subtract the band penalty from the logit of every
reflection token $u\in\mathcal{R}$ at each position $t$,
\begin{equation}
\tilde z_t(u)\;=\;
\begin{cases}
z_t(u)-\beta_i, & u\in\mathcal{R},\\[2pt]
z_t(u),         & u\notin\mathcal{R},
\end{cases}
\label{eq:penalty}
\end{equation}
then sample from $\mathrm{softmax}(\tilde z_t)$. The penalty is monotone in the reward: a
low-reward step reflects and explores freely ($\beta_i=0$), a mid-reward step is mildly
discouraged, and a high-reward step, likely on a reliable track, is strongly suppressed so
the model drives toward its conclusion instead of re-opening branches. Acting purely on
logits, this adds no tokens and no extra forward pass.

\paragraph{Early stopping via answer confidence.}
Additional reasoning helps only while the model is uncertain; once it is confident, further
steps burn tokens without changing an answer it has settled on. A persistent high-reward
state signals such a trajectory, but reward alone cannot decide whether to \emph{commit}: it
scores each reasoning \emph{step}, not confidence in the \emph{final answer}. We therefore add
a confidence probe: once the reward stays in the top band ($v_i\ge\tau_h$) for two consecutive
steps, we insert a closing prompt (\emph{``Okay, I think I have finished thinking.''}) that
exits the thinking phase and elicits a tentative answer $a=(x_1,\dots,x_n)$, whose perplexity
\begin{equation}
\mathrm{PPL}(a)
=\exp\!\left(-\frac{1}{n}\sum_{t=1}^{n}\log p(x_t\mid x_{<t})\right)
\label{eq:ppl}
\end{equation}
measures how confident the model is in that answer. We stop and commit when
\begin{equation}
\mathrm{PPL}(a)\le\tau_p,
\label{eq:stop}
\end{equation}
and otherwise discard the probe and resume reasoning. Reward and perplexity are
complementary, reward certifies that the \emph{trajectory} is sound, while perplexity gauges
confidence in the \emph{answer}, and coupling them helps in two ways a perplexity threshold alone cannot.
It suppresses \emph{confident-but-wrong} stops, since a fluent answer can have low perplexity
even when the reasoning never converged; and it confines the probe, which costs an extra
generation, to already-promising trajectories rather than running it at every step.

\begin{table*}[t]
\centering
\small
\resizebox{\textwidth}{!}{
\begin{tabular}{lccccccc ccccccc cc}
\toprule
\multirow{2}{*}{\textbf{Method}} & \multicolumn{7}{c}{\textbf{Accuracy (\%)~$\uparrow$}} & \multicolumn{7}{c}{\textbf{Average Tokens~$\downarrow$}} & \multirow{2}{1.1cm}{\centering\textbf{Latency (s)}~$\downarrow$} & \multirow{2}{*}{\textbf{Speedup}~$\uparrow$} \\
\cmidrule(lr){2-8}\cmidrule(lr){9-15}
 & GSM8K & MATH500 & AMC & AIME24 & AIME25 & GPQA & Avg. & GSM8K & MATH500 & AMC & AIME24 & AIME25 & GPQA & Avg. & & \\
\midrule
\multicolumn{17}{l}{\textit{DeepSeek-R1-Distill-Llama-8B}} \\
\midrule
Full CoT & 89.8 & 83.8 & 82.5 & 46.7 & 40.0 & 33.8 & 62.8 & 1201.2 & 4409.8 & 4986.6 & 11169.7 & 9329.3 & 11372.5 & 7078.2 & 56.01 & 1.00$\times$ \\
SnapKV   & 87.0 & 59.6 & 17.5 & 13.3 & 20.0 & 27.3 & 37.5 & 1696.8 & 7584.7 & 14695.7 & 14211.4 & 15216.1 & 14189.8 & 11265.8 & 49.07 & 1.14$\times$ \\
R-KV     & 88.9 & 61.0 & 65.0 & 26.7 & 16.7 & 30.3 & 48.1 & 1551.3 & 7891.7 & 8098.3 & 13227.6 & 13091.4 & 14933.9 & 9799.0 & 51.56 & 1.09$\times$ \\
RPC      & 90.5 & 80.4 & 70.0 & 20.0 & 16.7 & 38.9 & 52.8 & 1262.6 & 5627.7 & 8443.0 & 15074.7 & 14259.4 & 11551.6 & 9369.8 & 41.99 & 1.33$\times$ \\
SAT      & 87.1 & 79.2 & 75.0 & 40.0 & 36.7 & 30.8 & 58.1 & \textbf{248.6} & \textbf{2063.3} & 3724.3 & 8161.8 & \textbf{6354.1} & 5038.0 & \textbf{4265.0} & 28.29 & 1.98$\times$ \\
Dynasor  & 82.3 & 80.2 & 80.0 & 33.3 & 30.0 & 40.4 & 57.7 & 688.5 & 2615.7 & 4029.6 & 8788.2 & 9806.5 & \textbf{3869.5} & 4966.3 & 30.18 & 1.86$\times$ \\
\textbf{ReCo (Ours)} & 89.9 & 80.6 & 80.0 & 43.3 & 33.3 & 33.8 & \textbf{60.2} & 765.1 & 2713.9 & \textbf{2692.6} & \textbf{5345.3} & 8551.6 & 6875.7 & 4490.7 & \textbf{26.88} & \textbf{2.08$\times$} \\
\midrule
\multicolumn{17}{l}{\textit{DeepSeek-R1-Distill-Qwen-7B}} \\
\midrule
Full CoT & 88.1 & 87.8 & 80.0 & 43.3 & 33.3 & 38.9 & 61.9 & 1067.6 & 3268.7 & 4293.6 & 9783.7 & 8168.2 & 8939.6 & 5920.2 & 52.58 & 1.00$\times$ \\
SnapKV   & 87.7 & 76.0 & 65.0 & 23.3 & 13.3 & 44.4 & 51.6 & 1863.9 & 3987.9 & 14070.2 & 8945.8 & 14853.6 & 9644.6 & 8894.3 & 47.36 & 1.11$\times$ \\
R-KV     & 87.5 & 89.0 & 60.0 & 20.0 & 16.7 & 26.8 & 50.0 & 2123.3 & 4538.1 & 7473.6 & 9751.2 & 12520.8 & 13415.6 & 8303.8 & 45.58 & 1.15$\times$ \\
RPC      & 83.4 & 67.6 & 60.0 & 20.0 & 23.3 & 43.9 & 49.7 & 2446.3 & 6817.6 & 9476.8 & 11400.1 & 12382.4 & 11711.5 & 9039.1 & 43.42 & 1.21$\times$ \\
SAT      & 79.5 & 80.2 & 77.5 & 53.3 & 20.0 & 36.9 & 57.9 & \textbf{188.0} & 2165.2 & 3424.6 & 8750.3 & 8692.1 & 3916.1 & 4522.7 & 23.54 & 2.23$\times$ \\
Dynasor  & 83.6 & 75.0 & 72.5 & 36.7 & 26.7 & 35.4 & 55.0 & 363.8 & \textbf{1403.0} & 2850.2 & 6680.9 & 8428.3 & \textbf{1347.9} & 3512.4 & 24.56 & 2.14$\times$ \\
\textbf{ReCo (Ours)} & 89.1 & 78.8 & 80.0 & 40.0 & 30.0 & 41.9 & \textbf{60.0} & 451.2 & 1484.8 & \textbf{2175.5} & \textbf{2997.5} & \textbf{3440.2} & 1853.5 & \textbf{2067.1} & \textbf{22.35} & \textbf{2.35$\times$} \\
\bottomrule
\end{tabular}
}
\caption{Main results on DeepSeek-R1-Distill-Llama-8B and DeepSeek-R1-Distill-Qwen-7B
across six benchmarks. All reported numbers are averaged over three independent runs.
``Avg.'' averages over the six datasets. Latency is per-problem
end-to-end wall-clock time (s) measured on a single NVIDIA H20 GPU, including Pilot scoring
and the confidence probe for ReCo; speedup is relative to Full CoT.}
\label{tab:main}
\end{table*}
\begin{table}[t]
\centering
\setlength{\tabcolsep}{3pt}
\resizebox{\columnwidth}{!}{
\begin{tabular}{lcccc cccc cc}
\toprule
\multirow{2}{*}{\textbf{Method}} & \multicolumn{4}{c}{\textbf{Accuracy (\%)~↑}} & \multicolumn{4}{c}{\textbf{Avg. Tokens~↓}} & \multirow{2}{*}{\textbf{Lat.}~↓} & \multirow{2}{*}{\textbf{Sp.}~↑} \\
\cmidrule(lr){2-5}\cmidrule(lr){6-9}
 & GSM8K & GPQA & AIME24 & Avg. & GSM8K & GPQA & AIME24 & Avg. & & \\
\midrule
\multicolumn{11}{l}{\textit{Qwen3-8B}} \\
\midrule
Full CoT & 96.1 & 64.1 & 56.7 & 72.3 & 1268 & 8164 & 10932 & 6788 & 114.89 & 1.00x \\
SnapKV   & 92.4 & 39.9 & 23.3 & 51.9 & 1960 & 12384 & 14217 & 9520 & 70.69 & 1.63x \\
R-KV     & 93.4 & 46.5 & 30.0 & 56.6 & 2237 & 13878 & 14999 & 10371 & 104.89 & 1.10x \\
RPC      & 91.9 & 42.9 & \textbf{63.3} & 66.0 & 2479 & 14690 & 12891 & 10020 & 96.03 & 1.20x \\
SAT      & 93.0 & 52.0 & 56.7 & 67.2 & \textbf{885} & 5646 & 9496 & 5343 & 64.99 & 1.77x \\
Dynasor  & 93.3 & 59.6 & 50.0 & 67.6 & 1114 & \textbf{4037} & 9230 & 4794 & 60.67 & 1.89x \\
\textbf{ReCo} & \textbf{95.3} & \textbf{60.1} & 53.3 & \textbf{69.6} & 1046 & 5382 & \textbf{4530} & \textbf{3652} & \textbf{52.60} & \textbf{2.18x} \\
\bottomrule
\end{tabular}
}
\caption{Results on Qwen3-8B. Metrics follow Table~\ref{tab:main}.}
\label{tab:qwen3_8b}
\end{table}
\section{Experiments}
\subsection{Experimental Setup}
\label{sec:setup}

\paragraph{Models and datasets.}
We evaluate on three reasoning models across two families and scales:
DeepSeek-R1-Distill-Qwen-7B, DeepSeek-R1-Distill-Llama-8B~\cite{guo2025deepseek},
and Qwen3-8B~\cite{yang2025qwen3}. We benchmark on six datasets from grade-school
arithmetic to competition level: five math datasets, GSM8K~\cite{cobbe2021training},
MATH-500~\cite{hendrycks2021measuring}, AMC2023, AIME24, AIME25, and the scientific
GPQA~\cite{rein2023gpqa}.

\paragraph{Baselines.}
We compare against representative methods from both sides of reasoning efficiency.
\emph{(i)} Full CoT, the uncompressed model that keeps the full KV cache without length
control, serving as the accuracy reference.
\emph{(ii)} KV-cache compression: SnapKV~\cite{li2024snapkv}, R-KV~\cite{cai2026r},
and RPC~\cite{song2026reasoning}, which evict cache entries but do not control generation
length.
\emph{(iii)} Generation-length and early-exit methods: SAT~\cite{huang2026sat}, which
shortens reasoning via prompt-based control, and Dynasor~\cite{fu2025reasoning}, which
early-stops decoding; both keep the full cache.
Each addressing only the cache or only the generation, these baselines let us isolate
the benefit of our \emph{joint, reward-coordinated} design.

\paragraph{Implementation details.}
All experiments are conducted on a single NVIDIA H20 GPU. For KV-cache compression, we set the
base retention ratio to \(\lambda=0.25\), the reward-adjustment range to
\(\delta=0.1\), and the attention window to \(w=32\);
all KV-cache baselines use a comparable ratio of \(0.25\). For generation control, we
set the state thresholds to \(\tau_{\ell}=0.4\) and \(\tau_{h}=0.8\),
and the early-stopping perplexity threshold to \(\tau_p=1.1\). For a fair comparison, SAT and Dynasor are tuned to a comparable level of acceleration.
\subsection{Main Results}

Tables~\ref{tab:main} and~\ref{tab:qwen3_8b} report accuracy, average generated tokens,
and end-to-end latency across three models. We highlight three observations, one per metric.

\paragraph{ReCo best preserves accuracy under compression.}
ReCo stays closest to Full CoT: its average accuracy is $60.2\%$ vs.\ $62.8\%$ on Llama-8B,
$60.0\%$ vs.\ $61.9\%$ on Qwen-7B, and $69.6\%$ vs.\ $72.3\%$ on Qwen3-8B, the smallest gap
among all compressed methods. KV compression baselines fall much further behind (e.g.\ SnapKV
to $37.5\%$ and R-KV to $48.1\%$ on Llama-8B), and the gap widens on the hardest benchmarks:
on AIME25, ReCo retains $33.3\%$ on Llama-8B while every KV baseline collapses to $\le 20\%$.
Compressing the cache without accounting for the role of each step disrupts the context most
at the fragile, information-dense states that complex reasoning depends on, whereas ReCo's
reward-guided retention holds back compression exactly when the state is fragile.

\paragraph{ReCo reduces tokens without inflation, unlike cache-only compression.}
ReCo uses $37\%$, $65\%$, and $46\%$ fewer tokens than Full CoT on Llama-8B, Qwen-7B, and
Qwen3-8B, attaining the shortest average length on the latter two. In sharp contrast, the
cache-only methods SnapKV, R-KV, and RPC all \emph{increase} token count (e.g.\
$7{,}078\!\to\!11{,}266$ under SnapKV on Llama-8B), directly corroborating the length-inflation
effect of Sec.~\ref{subsec:motivation_length}. Length-only methods (SAT, Dynasor) can emit even
fewer tokens on some sets, but only by sacrificing accuracy (e.g.\ Dynasor drops to $82.3\%$ on
Llama-8B GSM8K, below Full CoT's $89.8\%$). Coupling compression with generation control lets
ReCo cut tokens aggressively while staying close to Full CoT accuracy.

\paragraph{ReCo delivers strong speedups without trading away accuracy.}
Reducing both the per-token attention cost and the generation length, ReCo
reaches $2.08\times$, $2.35\times$, and $2.18\times$ speedup over Full CoT on Llama-8B,
Qwen-7B, and Qwen3-8B. Cache-only baselines gain only modest speedups
($1.09$--$1.33\times$ on Llama-8B), since their inflated generation cancels much of the
per-token attention saving. Length-only methods (SAT, Dynasor), tuned to a
comparable level of acceleration, reach similar latency but only by trading away accuracy
(e.g.\ SAT at $58.1\%$ vs.\ ReCo at $60.2\%$ on Llama-8B). These results validate
our central claim: coordinating cache compression and generation control under a single reward
reduces both token count and latency while preserving reasoning quality, a balance that
handling either alone cannot achieve.\footnote{ReCo's own components add only a small
fraction of end-to-end inference time: Pilot scoring $0.60\%$, KV compression $0.11\%$,
the reflection penalty $1.24\%$, and early stopping $1.98\%$ on average.}
\subsection{Ablation Studies and Analysis}

\begin{figure}[ht]
\centering
\includegraphics[width=\columnwidth]{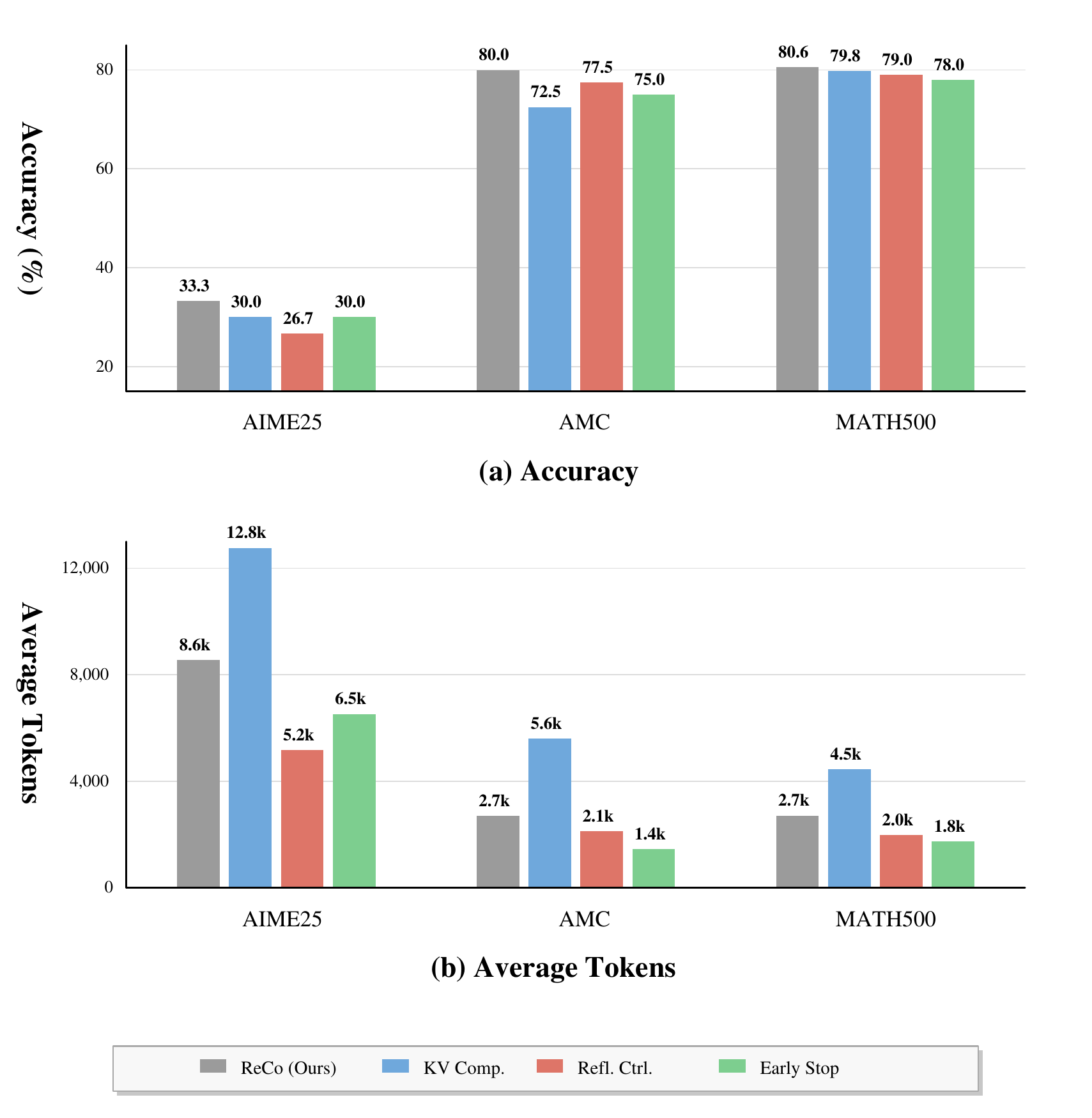}
\caption{Component ablation on DeepSeek-R1-Distill-Llama-8B: (a) accuracy (\%) and
(b) average tokens.}
\label{fig:ablation}
\end{figure}
\paragraph{The three components are complementary.}
Figure~\ref{fig:ablation} keeps only one component at a time. No single variant matches
ReCo's accuracy--cost balance. ``KV Comp.'' alone leaves generation unconstrained and is
the most expensive on hard problems ($12.8$k tokens on AIME25 vs.\ ReCo's $8.6$k), echoing
the length inflation of Sec.~\ref{subsec:motivation_length}. ``Refl.\ Ctrl.'' and ``Early
Stop'' curb tokens but, lacking reward-guided retention, lose accuracy on the harder sets.
ReCo is best on all three, AIME25 ($33.3\%$), AMC ($80.0\%$), MATH500 ($80.6\%$), showing
the components are complementary, not redundant. Intuitively, reward-guided compression
decides \emph{what} to keep, while reflection control and early stopping decide \emph{how far}
to reason; removing either half breaks the accuracy--cost trade-off that the full framework
maintains.

\begin{figure}[!ht]
\centering
\includegraphics[width=\columnwidth]{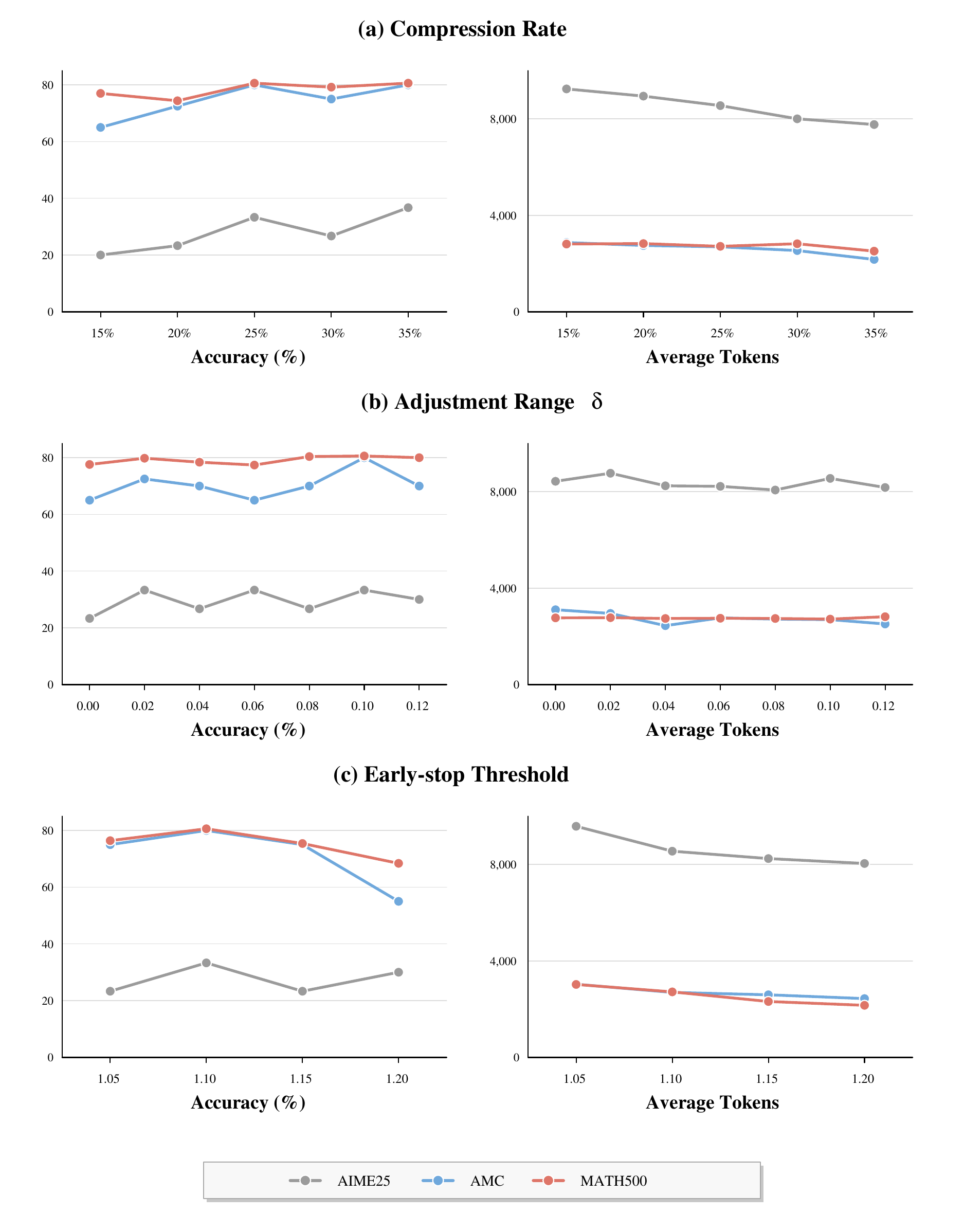}
\caption{Hyper-parameter sensitivity of ReCo (Llama-8B).}
\label{fig:hyperparam_ablation}
\end{figure}

\paragraph{Sensitivity to the base retention ratio.}
Row (a) of Figure~\ref{fig:hyperparam_ablation} sweeps the base retention ratio from
$15\%$ to $35\%$ ($\delta=0.10$). Aggressive pruning both destabilizes accuracy (at $15\%$,
AIME25 falls to $20.0\%$, AMC to $65.0\%$) and, as our analysis reveals, \emph{inflates} tokens
(AIME25 rises from $7.8$k at $35\%$ to $9.2$k at $15\%$), exactly the length-inflation failure
mode of Sec.~\ref{subsec:motivation_length}. Accuracy
stabilizes over $25$--$35\%$; we adopt the more aggressive $25\%$, which already delivers
high accuracy ($80.6\%$ MATH500, $80.0\%$ AMC) at the lowest per-token attention cost in
this stable range.

\paragraph{Sensitivity to the reward-adjustment range.}
Row (b) fixes $\lambda=0.25$ and varies $\delta$ from $0.00$ to $0.12$. At $\delta=0.00$
the policy is uniform and reward-agnostic, and is consistently among the weakest ($65.0\%$
on AMC), direct support for our motivation that reward-guided allocation beats treating all
steps alike. Accuracy peaks at $\delta=0.10$ ($80.0\%$ AMC, $80.6\%$ MATH500, $33.3\%$
AIME25), which we adopt as a robust default.

\paragraph{Sensitivity to the early-stopping threshold.}
Row (c) varies $\tau_p$ from $1.05$ to $1.20$, trading length for accuracy: a stricter
threshold preserves accuracy but emits more tokens, a looser one halts earlier at some
accuracy risk (MATH500 drops $80.6\%\!\to\!68.4\%$ from $1.10$ to $1.20$). $\tau_p\!=\!1.10$
is best on all three sets ($33.3\%$ AIME25, $80.0\%$ AMC, $80.6\%$ MATH500) while already
saving many tokens. We use one configuration ($\lambda=0.25$, $\delta=0.10$,
$\tau_p=1.10$) for all models and datasets, without per-task tuning.
\begin{table}[ht]
    \centering
    \small
    \setlength{\tabcolsep}{4pt}
    \begin{tabularx}{\columnwidth}{@{}l *{4}{>{\centering\arraybackslash}X}@{}}
    \toprule
    \textbf{Method} & Full-CoT & R-KV & SAT & \textbf{ReCo} \\
    \midrule
    \textbf{Max Memory (GB)} & 28.78 & 20.70 & 26.90 & \textbf{17.92} \\
    \bottomrule
    \end{tabularx}
    \caption{Peak GPU memory on AIME25 (Llama-8B).}
    \label{tab:memory}
    \end{table}
\paragraph{Peak memory usage.}
Table~\ref{tab:memory} reports the maximum GPU memory on AIME25 with
DeepSeek-R1-Distill-Llama-8B. ReCo uses $17.92$\,GB, a $37.7\%$ reduction over full cache
($28.78$\,GB) and below both R-KV ($20.7$\,GB) and SAT ($26.9$\,GB). A KV-only compressor
like R-KV still trails ReCo because its inflated outputs enlarge the very cache it tries to
shrink; by coupling compression with generation control, ReCo attains the lowest footprint.

\section{Related Work}

\noindent\textbf{Efficient Reasoning.}
Long CoT boosts LRMs but inflates token and compute
cost~\cite{feng2025efficient, qu2025survey, yue2025don}. Prior work spans length-aware RL,
fine-tuning, or distillation for compact reasoning~\cite{yeo2025demystifying, cheng2026optimizing, luo2026ada, zeng2025done, ning2025not, tang2026towards, luo2026o1, aggarwal2025l1},
inference-time prompting or explicit token budgets~\cite{han2025token, huang2026sat, renze2024benefits, nayab2024concise},
early stopping~\cite{yang2025dynamic, qiao2025concise, wang2025wait}, and model
routing~\cite{ong2024routellm, chen2025tagrouter, song2025irt}. Yet these methods typically
retrain the reasoning model or touch only the generated text, leaving the KV cache untouched. ReCo needs no training of the reasoning model and
no per-task prompt engineering, only a $30$M step-level reward estimator, allocates per
step under a single process reward, and controls cache and length jointly.

\noindent\textbf{KV Cache Compression.}
The KV cache is a major latency and memory bottleneck in LLM
inference~\cite{li2024survey, shi2024keep},
addressed by token eviction~\cite{li2024snapkv, cai2024pyramidkv, feng2026ada, yang2026indexmem},
merging~\cite{zhang2024cam, wang2024model, wan2024d2o},
quantization~\cite{hooper2024kvquant, zandieh2025turboquant, han2025polarquant}, and low-rank
decomposition~\cite{chang2024palu, chang2025xkv}, but they assume a static prompt pruned
once after prefilling. Recently adapted to reasoning, where the cache instead grows with the
chain of thought, R-KV~\cite{cai2026r} and RPC~\cite{song2026reasoning} use heuristic
eviction that ignores step importance and acts on the cache alone, so length inflation
erodes the saving. ReCo
instead ranks retention by per-step process reward and couples compression with generation
control.

\section{Conclusion}
We presented ReCo, a step-wise framework that coordinates KV-cache compression with
generation control under a single per-step process reward. Our analysis shows that
reasoning states differ in how much compression they tolerate and that compressing the cache alone often
lengthens the subsequent reasoning, so the extra generation cancels much of the intended
saving. By governing the cache and the
reasoning length jointly, ReCo cuts token count and latency across three
reasoning models and six benchmarks, and lowers peak memory, while staying close to
full-cache accuracy, indicating that KV compression is better coupled with generation
control.

\bibliography{reco}

\end{document}